\documentclass{article} 
\usepackage{arxiv}

\usepackage[utf8]{inputenc} 
\usepackage[T1]{fontenc}    
\usepackage{hyperref}       
\usepackage{url}            
\usepackage{booktabs}       
\usepackage{amsfonts}       
\usepackage{nicefrac}       
\usepackage{microtype}      
\usepackage{lipsum}		
\usepackage{graphicx}
\usepackage{natbib}
\usepackage{doi}

\renewcommand{\shorttitle}{Robofriend TR}

\hypersetup{
pdftitle={Robofriend: An Adpative Storytelling Robotic Teddy Bear -- Technical Report},
pdfsubject={cs.Robotics},
pdfauthor={Ido Glanz, Matan Weksler, Erez Karpas, Tzipi Horowitz-Kraus},
pdfkeywords={Robotics, Education, Reinforcement Learning},
}

\usepackage{scalerel,amssymb}

\title{Robofriend: An Adpative Storytelling Robotic Teddy Bear -- Technical Report}
\author{
    Ido Glanz$^\ast$ \\
    Technion Autonomous Systems Program\\
    Technion -- Israel Institute of Technology\\
    \And 
    Matan Weksler$^\ast$ \\
    Technion Autonomous Systems Program\\
    Technion -- Israel Institute of Technology\\
    \And 
    Erez Karpas \\
    Faculty of Data and Decision Sciences\\
    Technion -- Israel Institute of Technology\\
    \texttt{karpase@technion.ac.il}\\
    \And
    Tzipi Horowitz-Kraus\\
    Faculty of Education in Science and Technology\\
    Technion -- Israel Institute of Technology\\
    \texttt{tzipi.kraus@ed.technion.ac.il}\\
}
\date{}

\usepackage{color}

\newcommand\blfootnote[1]{%
  \begingroup
  \renewcommand\thefootnote{}\footnote{#1}%
  \addtocounter{footnote}{-1}%
  \endgroup
}

\begin{document}

\maketitle
\blfootnote{$^\ast$ denotes equal contribution}

\begin{abstract}
In this paper we describe Robofriend, a robotic teddy bear for telling stories to young children. Robofriend adapts its behavior to keep the childrens' attention using reinforcement learning.
\end{abstract}


\section{Introduction}

Language exposure at an early stage of development is critical for the facilitation of brain networks associated with language \cite{kuhl1,Cardillo1,moon1}. Storytelling is one form  of language exposure, which was found to be associated with a greater engagement not only in language processing but also in visualization and cognitive abilities in children \cite{hutton1}. Interestingly, it was suggested that it is not the storytelling itself that is related to these improvements, but it is the interaction during the stories that amplify these abilities in children \cite{twait1}. A recent study demonstrated how a group of 4--6-year-old children attending storytelling sessions interactively vs. a group attending non-interactively (storytelling sessions on the screen), shared greater cognitive and language abilities \cite{twait1}. Hence, a question was raised regarding this positive effect during interactive (dialogic) storytelling --  is the positive effect due to the human interaction? or due to the interactive nature of the storytelling? in other words, will an interactive robot during storytelling result in similar results as the human-based interactive condition? 

\begin{figure}
    \centering
    \includegraphics[width=\linewidth]{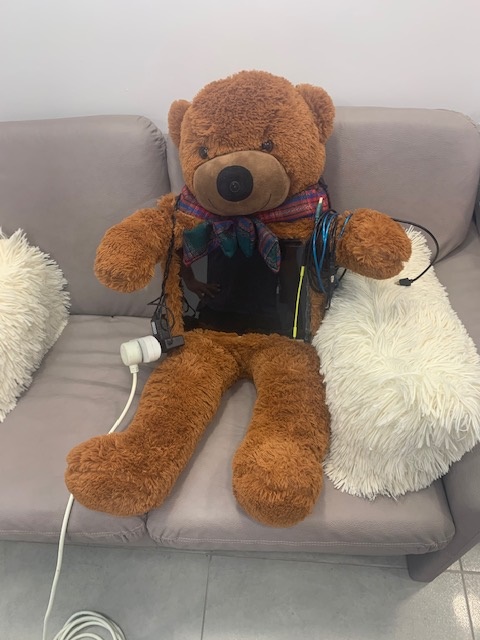}
    \caption{Robofriend in Home Testing}
    \label{f:robofriend_photo}
\end{figure}

To study this question, we designed Robofriend (shown in Figure \ref{f:robofriend_photo}) -- a robotic teddy bear that reads young children stories. 
Robofriend is constructed by taking a regular teddy bear and inserting a tablet in its belly, as well as a rudimentary skeleton, motors and sensors that allow it to move its head and arms. 
Robofriend can read a story to a small group of children, with the robot's main objective being to engage the children, keeping their attention on the story. Thus, although our main motivation for designing Robofriend is the scientific study described above, Robofriend can also serve as a tool that a teacher in a daycare class can use. Robofriend can read a story to one group of children while the teacher engages with the other children in the class.

Each story that Robofriend can tell is divided into prerecorded video segments. Typically, each segment will correspond to showing a still image of one page in the printed book, with a human reading the text on the page. Note that Robofriend does not perform any text to speech, the segments are all prerecorded. At the end of each segment, Robofriend chooses which action to perform out of several actions it has available. 
Possible actions include asking a simple question about the story (there is a set of prerecorded questions for each segment of the story, Robofriend chooses one of these randomly), giving positive feedback (e.g., ``very good children, I see you are paying attention''), or negative feedback (e.g., ``children, are you listening to the story?''). 

As previously stated, the objective of Robofriend is to keep the childrens' attention. The first step to optimizing something is to measure it, or at least some proxy of it. Robofriend uses a camera to measure some things, which can serve as a proxy for engagement. First, Robofriend uses computer vision to detect the faces of the children, and the direction of their gazes.
We remark that these faces are anonymous -- Robofriend does not try to associate faces to identities in any way. From these face detection, we extract several measurements: how many faces are looking at the robot, how focused are they on the story (using their relative gaze) as well as how ``jumpy'' the faces are (an "excitement" metric). Aside from the visual attributes, we also monitor the noise level as and its momentary change (its derivative) to serve as supportive metrics capturing the children's state. 
These are aggregated into a 
reward signal for each camera video frame, and aggregated throughout each story segment to produce a state and reward for each segment.

Having defined the 
rewards, we can now try to optimize our objective -- the total sum of rewards. Of course, we do not know in advance what is the right action to take after each story segment, nor do we have a model for how each action will affect the children's engagement. Therefore, we chose to use reinforcement learning to control Robofriend's actions. However, because young children are involved, we do not want to allow the robot to explore sequences of actions we know are not beneficial for the children (for example, always using the negative feedback action). Therefore, we adopt the approach of using LTL ``restraining bolts'' \cite{DBLP:conf/aaai/GiacomoIFP20}, and manually encode what are the allowed trajectories for Robofriend.

In the remainder of this paper, we describe the design of Robofriend in more detail. We also describe our preliminary evaluation of the robot at a local daycare center. Finally, we conclude with some lessons learned and a discussion of the ethical considerations that arose in this project.

\section{Robofriend Design}

We now describe the design of Robofriend in more detail, starting with the mechanical build.

\subsection{Robofriend Mechanical Build}

\begin{table}
\small
    \centering
    \begin{tabular}{l|l|l|r}
    \hline
Item & Use & Quantity & Price\\
\hline
Bear doll & - & 1 & US\$30.00\\
Display & Play video & 1 & US\$200.00\\
Camera & Monitor kids & 1 & US\$79.00\\
Arduino + wiring kit & Control Servos & 1 & US\$37.00\\
Servo & Move head & 4 & US\$24.00\\
Speakers & Playing Sound & 1 & US\$20.00\\
\hline
Total: &  &  & US\$462.00\\
\hline
    \end{tabular}
    \caption{Robofriend Bill of Materials}
\label{t:bom}
\end{table}

\begin{figure}
    \centering
    \includegraphics[width=\linewidth]{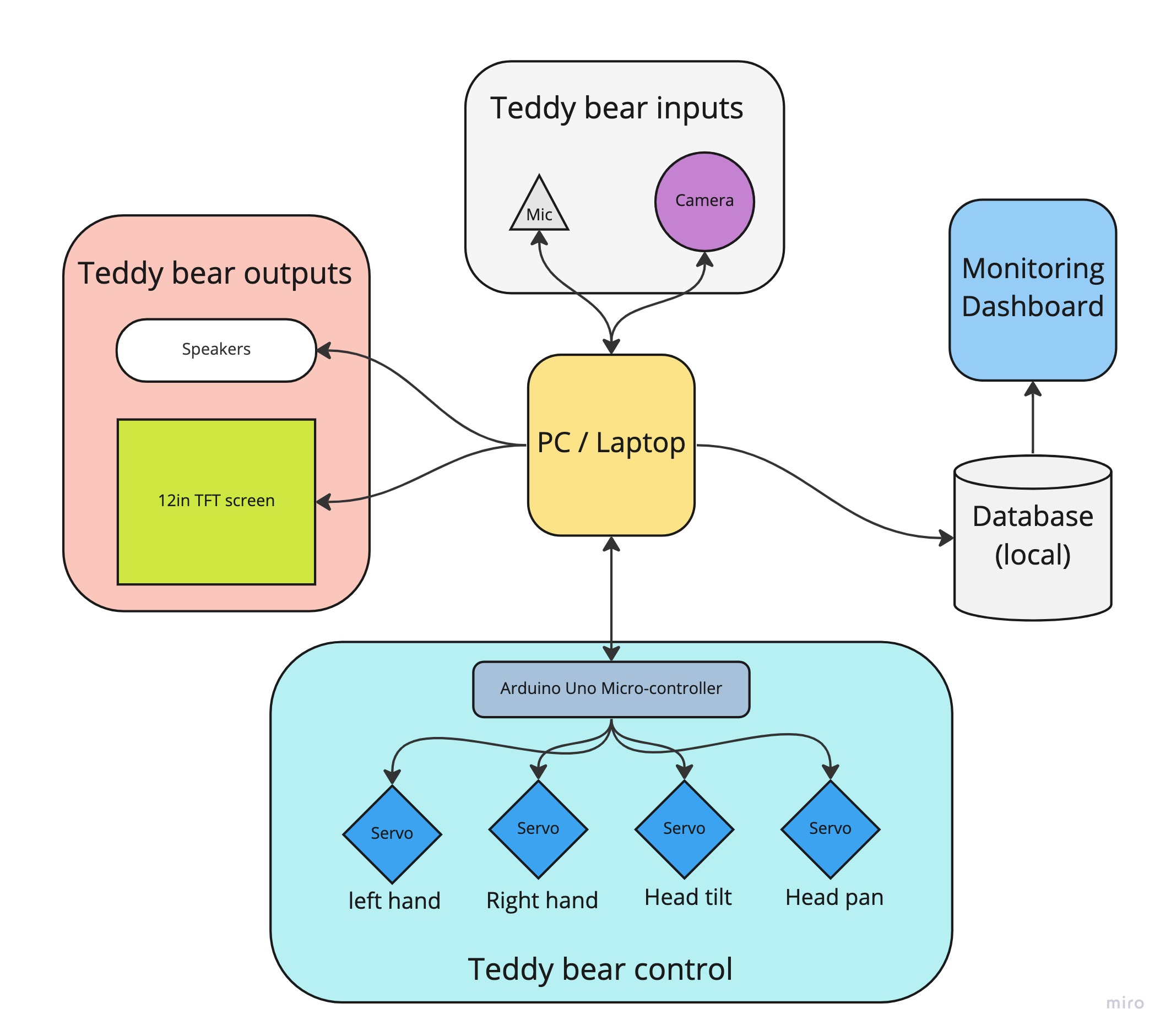}
    \caption{Hardware schematic block diagram}
    \label{f:hardware}
\end{figure}

As previously stated Robofriend is constructed by taking a large, 1m tall, teddy bear, and instrumenting it to be able to move its head and arms, play videos and sound, and look at the children it is reading the story to. First, we inserted an aluminum skeleton into the robot to support the other devices. To do so, we removed the majority of the stuffing and decoupled the head momentarily to mount the camera in the bear's nose and create a fixture for the servo motors to connect to. We then mounted a 12.3 inch display, which the robot displays the story segments on, as well as 4 servo motors which are used to move the head and arms (one for each arm and 2 for the the head pan and tilt). As mentioned above, a camera was placed into the teddy bear's nose, to monitor the children and measure the reward signal, finally, speakers were connected to play sound, see figure \ref{f:hardware} for a schematic diagram.

\begin{figure}
    \centering
    \includegraphics[width=\linewidth]{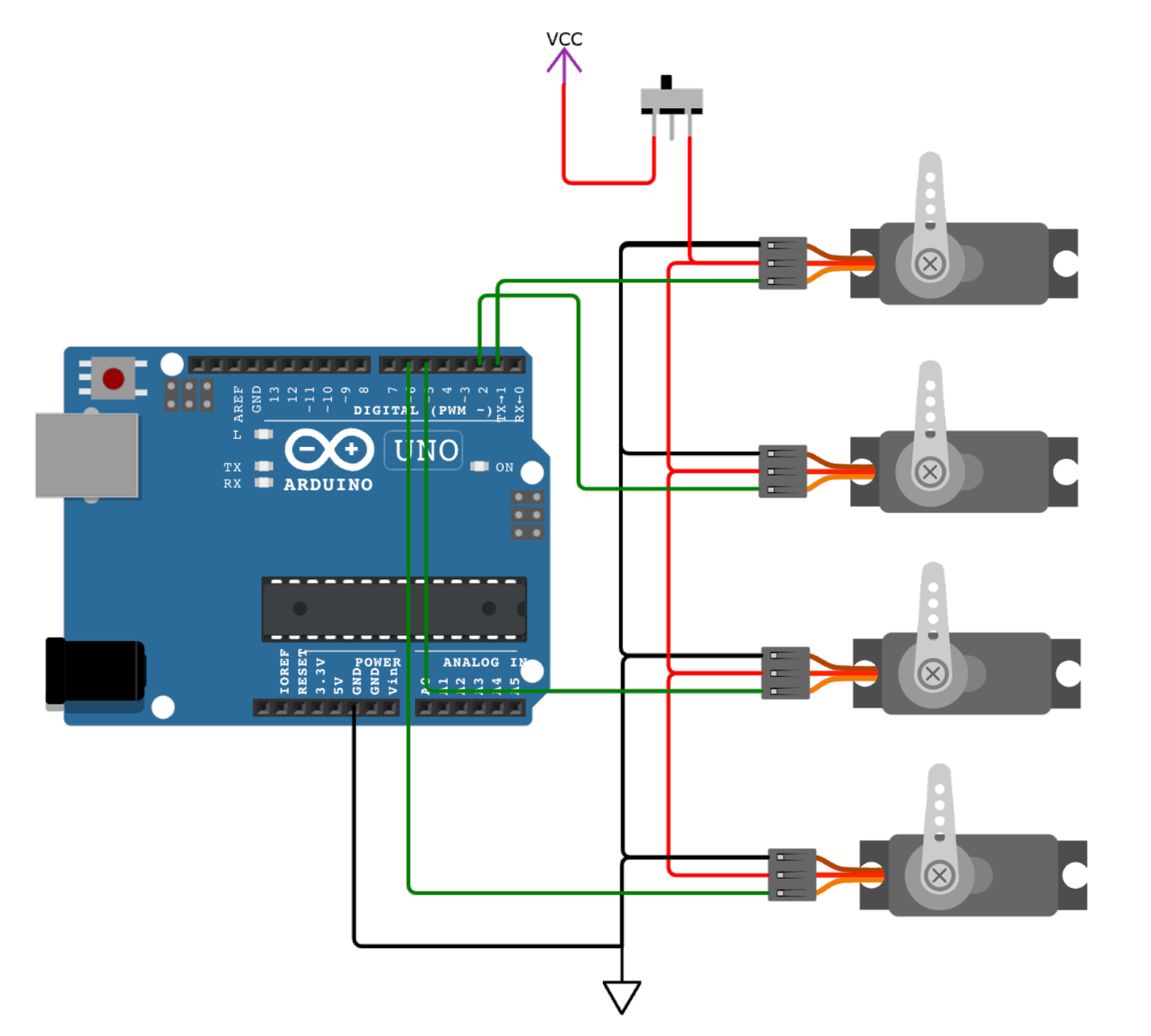}
    \caption{Arduino sub-system}
    \label{f:arduino}
\end{figure}

This hardware was controlled from a PC, which was connected directly to the camera, display, and speakers. The servo motors are controlled by an Arduino Uno (see Figure \ref{f:arduino}), which was connected to the PC as well. The controller for the servo motors runs in a separate process on the Arduino, following instructions from the PC.


The Bill of Materials (BOM) for Robofriend is shown in Table \ref{t:bom}, while a detailed BOM with links to each item is available online at: \url{shorturl.at/efBZ2}. 
Overall, the total cost to construct Robofriend was less than US\$500, making it fairly accessible.

\subsection{Robofriend Code Architecture}
To operate and coordinate the different algorithms, hardware and user interactions, a proprietary python software stack was developed and will be briefly described below. The code is available at \url{https://github.com/IdoMatan/RoboFriend}.

The code architecture is based on RabbitMQ, which is a ROS-like publisher-subscriber framework that implements  asynchronous parallel process control. RabbitMQ allows us to simultaneously control the robot's servos, camera, screen and any other needed peripherals, as well as to run the algorithms we will describe later.

\begin{figure}
    \centering
    \includegraphics[width=\linewidth]{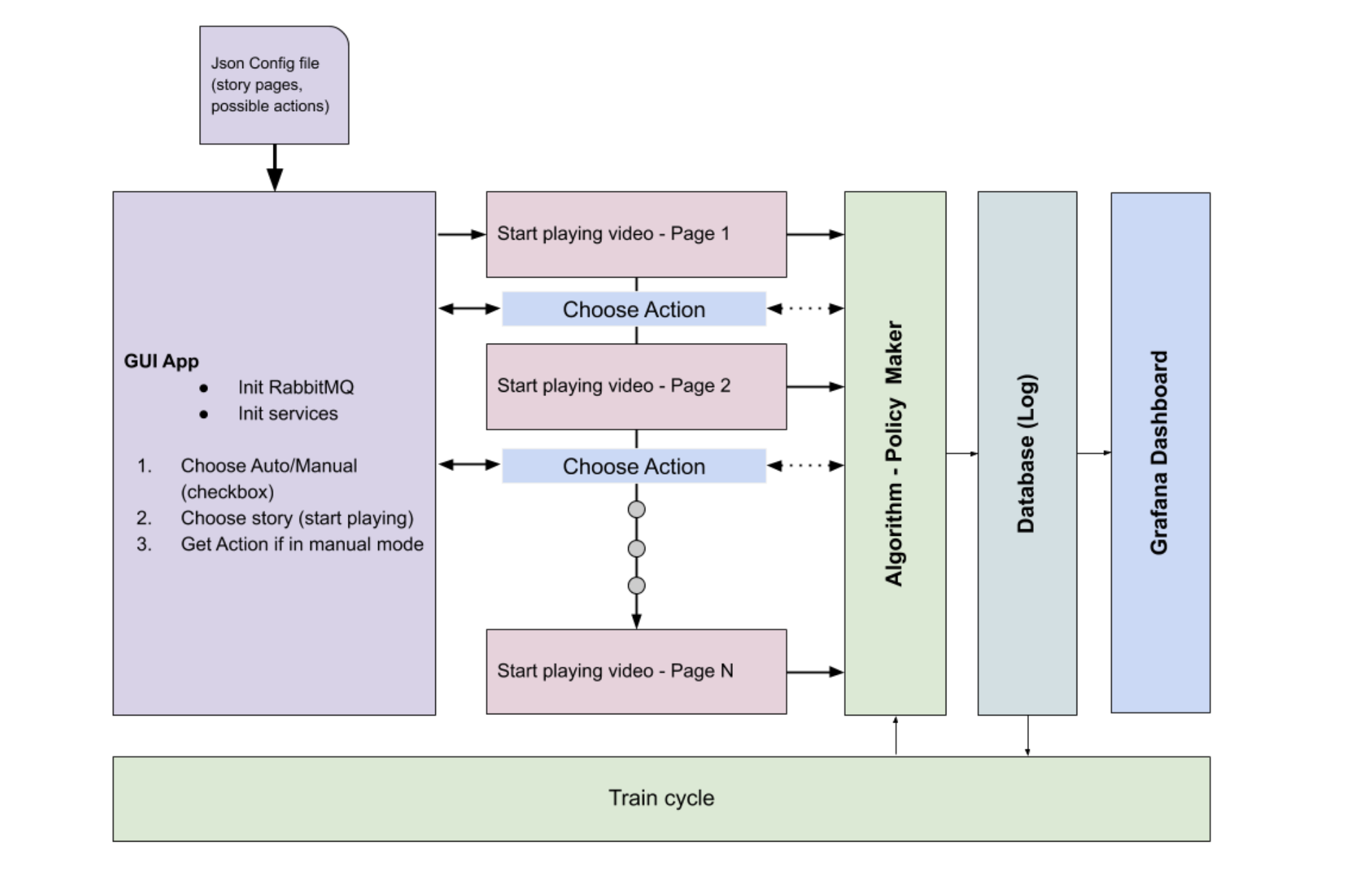}
    \caption{Software Operational Flow}
    \label{f:ops_schema}
\end{figure}

\begin{figure}
    \centering
    \includegraphics[width=\linewidth]{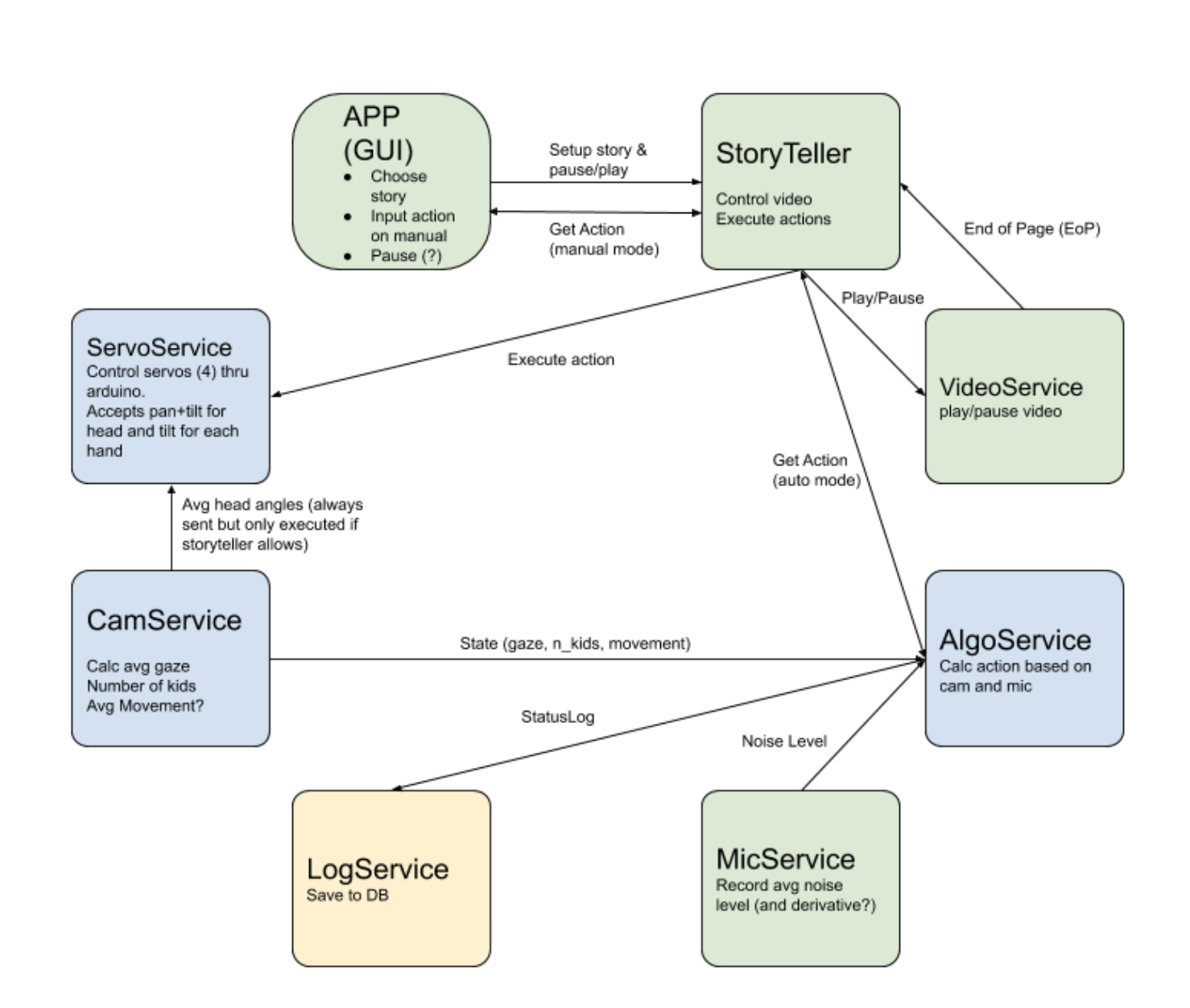}
    \caption{Software Architecture Diagram}
    \label{f:software1}
\end{figure}

Figure \ref{f:software1} shows the schematic structure of our software, showing the processes and the messages that are passed between them. 
The main process is the StoryTeller, which coordinates the flow among the other processes and displays the video.
This process runs a loop which plays the next story segment, then calls the algorithm service to get the next action. This loop repeats until the story ends. Throughout this loop, the robot moves its head, aiming to center its viewing angle so to center all faces in the frame.
This flow is illustrated in 
Figure \ref{f:ops_schema}.

The other services are either timer-based (e.g. send a frame every $N$ milliseconds) or event-driven, e.g., a page-ended message would trigger a next action calculation in the algorithm-service. 
All metrics, actions and useful metadata (not video footage) were logged in real-time in a local Postgres database, allowing both post-analysis of the trial as well as live monitoring using a Grafana dashboard.

\begin{figure}[t]
    \centering
    \includegraphics[width=\linewidth]{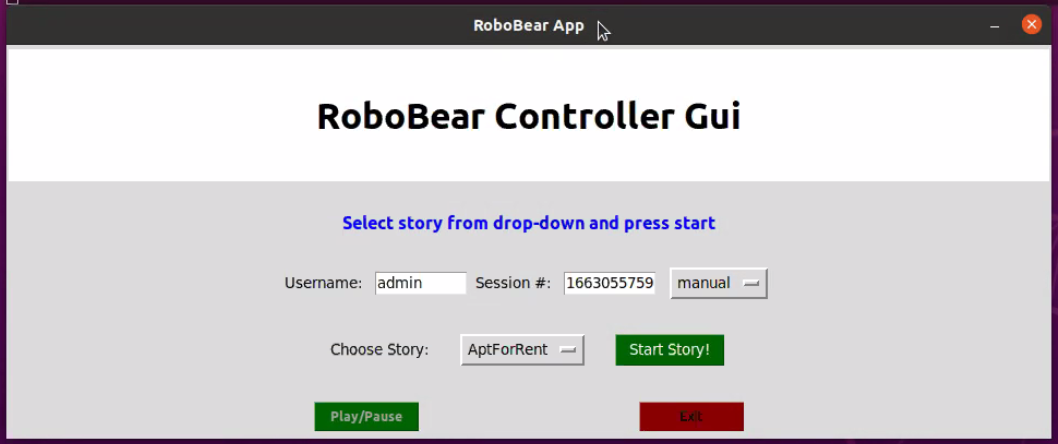}
    \caption{A simple python-based graphical user interface to interact with the robot}
    \label{f:gui}
\end{figure}

Figure \ref{f:gui} shows the simple GUI implemented where a user can run the app, choose one of the supported stories, an operation mode (which will be discussed later) and start and stop the story.

\subsection{Robofriend Algorithm}
Having described the mechanical construction of Robofriend and the code architecture, we can now discuss the algorithm which is used to control it. 
As previously mentioned, our high-level control algorithm involved using reinforcement learning with ``restraining bolts'' 
\cite{DBLP:conf/aaai/GiacomoIFP20}, to avoid the robot following trajectories which we know will not be good.
We begin by describing our sensing and reward function, then we describe the actions which are available to the robot, and finally, we describe the constraints which were used as the ``restraining bolts'' in our preliminary evaluation.

\subsubsection{Robofriend Sensing and Reward Function}

As previously mentioned, the reward is based on using computer vision to detect the children's faces and gaze direction, and on measuring the noise level.
Specifically, we used an MTCNN neural-network-based face detector to detect the faces of the children within the frame \cite{zhang2016joint}, followed by a gaze estimation step for each using GazeNet \cite{zemblys2018gazeNet} to generate a gaze vector relative to the camera lens.
This results in the 
following metrics:
\begin{description}
    \item[Number of Faces] The number of detected faces by the MTCNN face detection algorithm. A change in the number of faces would likely indicate a child walking away or not looking at the camera.
    
    \item[Average relative gaze (attention)] For each detected face (denoted by index $i$), a gaze vector $\theta_i$, $\phi_i$ is predicted by the GazeNet algorithm, where $\theta_i$ is the lateral (left/right) angle and $\phi_i$ is the vertical (up/down angle) -- both angles are relative to the center of the frame (the camera lens center). 
    Based on these measurements, we define the {\em gaze} component of the reward as:    
    $$
    r_{gaze} := \frac{1}{n}\sum_{i=1}^n cos(\theta_i)
    $$
    Roughly speaking, our attention metric, ranging from 0-1, corresponds to how focused the children are on the robot as opposed to looking around the room.
    
    \item[Excitement] Using consecutive frames we are able to calculate a per-face {\em jumpiness} metric corresponding to how still the children are. This is another proxy to their attention and engagement.
    Formally, let us denote the positions of (centers of) faces detected in the first image by $x_i, y_i$ (for $i=1 \ldots n$), and the positions of the faces detected in the next image by $x'_j, y'_j$ (for $j=1 \ldots m$). As the faces do not have identities associated with them, we must first align the faces in the first image to the faces in the next image. We do so greedily by finding, for each face position $x_i, y_i$ in the first image, the closest face position $x'_j, y'_j$ (using Euclidean distance) in the next image (which is also under a feasible max possible distance). We then define the jumpiness for face $i$ by this Euclidean distance, and the total jumpiness is the sum of jumpiness for each face, that is:
    $$
    r_{jump} := \sum_{i=1}^n \min_j \sqrt{(x_i - x'_j)^2 + (y_i - y'_j)^2 }
    $$
    \item[Noise Level] The average noise level over a 1 second period as measured by the microphone inside the teddy bear. This is a proxy for how much the children are talking to each other instead of listening to the story. We denote this by $r_{noise}$.

    \item[Deviation of Noise Level] The derivative of the noise level, again averaged every second, denoted by $r_{nd}$. The motivation behind this metric is to capture changes in the sound level within a page, indicating the children are getting noisier or quieter potentially due to the effect of the previous action.

\end{description}

To aggregate the reward signal throughout the duration of a story segment, we average each of these metrics for each video frame that belongs to this segment. Note that we use average instead of sum, as different story segments have different durations. This gives us a state vector $\langle r_{gaze}, r_{jump}, r_{noise}, r_{nd} \rangle$.

Finally, to aggregate these different metrics into a single reward function we use a weighed sum, and thus our reward denoted by $r$ is defined as:
$$
r := \alpha_1*r_{gaze} - \alpha_2*r_{jump} - \alpha_3*r_{noise} + \alpha_4*r_{nd} + ltl\_reward
$$
Where the $ltl\_reward$ will be described in details below, but conceptually corresponds to a set of pre-defined restraining rules the robot should learn to obey.

We conclude the discussion of the reward by noting that these measurements serve as a proxy for the real reward (the children's attention), which we can not measure directly. 


\subsubsection{Robofriend States and Actions}

As we described above, Robofriend reads a story, which is divided into segments, and chooses which action to perform after every segment.
Note that the order between the story segments is linear, and so there is no choice with regard to which story segment to read next. The only choice is which action to perform after every story segment.

Robofriend's actions correspond to different types of feedback it can give the children and are divided into:
\begin{description}
\item[Positive Feedback] This action randomly chooses from a set of positive feedback sentences, such as ``great job'' or ``you are listening nicely''.
\item[Negative Feedback] This action chooses randomly from a set of (mildly) negative feedback sentences, such as ``please pay attention'' or ``please be quiet''.
\item[Question] This action chooses a random question relating to the story segment Robofriend just finished reading. Robofriend does not attempt to extract an answer but merely pauses for an appropriate amount of time.
\item[Continue] Continue immediately to the next story segment.
\item[Move head and arms] Execute a series of random head and arm movements for a few seconds.
\end{description}

Note that all of these actions are applicable regardless of where Robofriend is in the story (the question action changes as a function of it, but is always applicable). Furthermore, the only information Robofriend extracts from its sensors is the reward signal. Thus, there is no notion of state for where we are in the story, and the only state information is the state vector $\langle r_{gaze}, r_{jump}, r_{noise}, r_{nd} \rangle$ described above.
However, the LTL constraints used for the ``restraining bolts'' do carry their own notion of state, as we discuss next.


\subsubsection{Robofriend Restraining Bolt}

The final part of Robofriend's control architecture is the ``restraining bolt'' -- the LTL constraint which the robot needs to obey \cite{DBLP:conf/aaai/GiacomoIFP20}. That is, we specify LTL trajectory constraints over allowed and forbidden {\em trajectories} for the robot, where in this case a trajectory is a sequence of actions.
For example, we can specify the constraint to not ask two questions in a row as
$$G(ask\_question \rightarrow X \neg ask\_question) $$
(that is, if the current action is ask question, then the next action is not ask question). This ensures the robot will not ask questions constantly.

For the preliminary evaluation of the robot, we tested the following restraining bolts which felt sensible to include, where the overall constraint is a conjunction of the following:

\begin{description}
    \item[Do not ask two questions in a row]   $$G(ask\_question \rightarrow X \neg ask\_question)$$
    \item[Do not wave hands twice in a row]  $$G(wave\_hands \rightarrow X \neg wave\_hands)$$
    \item[Eventually ask a question] $$F(ask\_question)$$
    \item[Eventually wave hands] $$F(wave\_hands)$$
\end{description}




\subsubsection{Robofriend Learning Algorithm}

Robofriend can use reinforcement learning to find a policy which optimizes expected sum of rewards, while still having a high chance of conforming to the LTL constraints. Specifically, we use an Actor-Critic reinforcement learning algorithm \cite{konda1999actor} which is fed at each step the state of the children and predicts the next action towards maximizing the expected sum of rewards throughout the episode, accounting also for the LTL constraints modifying the rewards \cite{DBLP:conf/aaai/GiacomoIFP20}.

Bearing in mind these type of algorithms often need a massive amount of diverse enough training data to converge, and with the specific problem setting complexity and training availability, we added an option to conduct imitation learning of the policy network based on an external "wizard of oz" feedback given in real-time by a supervision/teacher. Essentially, during the first trials of the robot, a teacher could manually select an appropriate action, and the algorithm would use that as a labeled training set to conduct initial training of its policy network. This allows to both give a better baseline to start the learning process as well as avoid unwanted situations in the very beginning of the training process.

\section{Ethical Considerations}

During the construction of Robofriend, many ethical considerations came up. We now discuss these in detail.

The first consideration is privacy, which is a potential problem, especially when children are involved. Therefore, we designed Robofriend in a way that maximizes the privacy of the children. First, Robofriend does not keep any recorded images. All images are processed online to detect faces (without any attempt to associate faces with names -- which the robot does not have anyway). The only data Robofriend records involve a count of how many faces it detected in any given frame, as well as the direction of their gaze. 

Another potential concern is that Robofriend will replace human contact with a teacher. As previously mentioned, Robofriend is meant to be a tool to be used by a daycare teacher, so that part of the class can listen to a story, while the teacher engages more personally with the remaining children. This can help the teacher devote more individual time to children who experience challenges attending to stories with background distractions, or to supplement activities for those who experience challenges with social interaction. Thus, Robofriend serves as a supplement for the teaching staff, and not as a replacement for the teacher. 


\section{Conclusion}

We have described the construction and software architecture for Robofriend, a robotic story-telling teddy bear. In future work we intend to examine how children react to Robofriend, as well as its ability to learn to engage children in the story.

\bibliographystyle{unsrtnat}

\begin{thebibliography}{9}
\providecommand{\natexlab}[1]{#1}
\providecommand{\url}[1]{\texttt{#1}}
\expandafter\ifx\csname urlstyle\endcsname\relax
  \providecommand{\doi}[1]{doi: #1}\else
  \providecommand{\doi}{doi: \begingroup \urlstyle{rm}\Url}\fi

\bibitem[Kuhl(2004)]{kuhl1}
P.~K. Kuhl.
\newblock Early language acquisition: cracking the speech code.
\newblock \emph{Nat Rev Neurosci}, 5\penalty0 (11):\penalty0 831--843, 2004.

\bibitem[Cardillo and Kuhl(2009)]{Cardillo1}
Lebdeva~GC Cardillo and P.~K. Kuhl.
\newblock Individual differences in infant speech perception predict language
  and pre-reading skills through age 5.
\newblock In \emph{Annual Meeting of the Society for Developmental and
  Behavioral Pediatrics}, 2009.

\bibitem[Moon et~al.(2013)Moon, Lagercrantz, and Kuhl]{moon1}
C~Moon, H.~Lagercrantz, and P.~K. Kuhl.
\newblock Language experienced in utero affects vowel perception after birth: a
  two-country study.
\newblock \emph{ACTA Paediatra}, 102\penalty0 (2):\penalty0 156--160, 2013.

\bibitem[Hutton et~al.(2015)Hutton, Horowitz-Kraus, Mendelsohn, DeWitt, and
  Holland]{hutton1}
JS~Hutton, T~Horowitz-Kraus, AL~Mendelsohn, T~DeWitt, and SK~Holland.
\newblock Home reading environment and brain activation in preschool children
  listening to stories.
\newblock \emph{Pediatrics}, 136\penalty0 (3):\penalty0 466--478, 2015.

\bibitem[Twait et~al.(2019)Twait, Farah, Shamir, and Horowitz-Kraus]{twait1}
E~Twait, R~Farah, N.~Shamir, and T.~Horowitz-Kraus.
\newblock Dialogic reading intervention in preschoolers is related to greater
  cognitive control: an eeg study.
\newblock \emph{ACTA Paediatra}, 108\penalty0 (11):\penalty0 1993--2000, 2019.

\bibitem[Giacomo et~al.(2020)Giacomo, Iocchi, Favorito, and
  Patrizi]{DBLP:conf/aaai/GiacomoIFP20}
Giuseppe~De Giacomo, Luca Iocchi, Marco Favorito, and Fabio Patrizi.
\newblock Restraining bolts for reinforcement learning agents.
\newblock In \emph{{AAAI}}, pages 13659--13662. {AAAI} Press, 2020.

\bibitem[Zhang et~al.(2016)Zhang, Zhang, Li, and Qiao]{zhang2016joint}
Kaipeng Zhang, Zhanpeng Zhang, Zhifeng Li, and Yu~Qiao.
\newblock Joint face detection and alignment using multitask cascaded
  convolutional networks.
\newblock \emph{IEEE signal processing letters}, 23\penalty0 (10):\penalty0
  1499--1503, 2016.

\bibitem[Zemblys et~al.(2018)Zemblys, Niehorster, and
  Holmqvist]{zemblys2018gazeNet}
Raimondas Zemblys, Diederick~C Niehorster, and Kenneth Holmqvist.
\newblock gazenet: End-to-end eye-movement event detection with deep neural
  networks.
\newblock \emph{Behavior research methods}, 2018.

\bibitem[Konda and Tsitsiklis(1999)]{konda1999actor}
Vijay Konda and John Tsitsiklis.
\newblock Actor-critic algorithms.
\newblock \emph{Advances in neural information processing systems}, 12, 1999.

\end{thebibliography}

\end{document}